\documentclass[10pt,twocolumn,letterpaper]{article}

\usepackage{dicta}
\usepackage{times}
\usepackage{epsfig}
\usepackage{graphicx}
\usepackage{amsmath}
\usepackage{amssymb}
\usepackage{booktabs}

\usepackage[pagebackref=true,breaklinks=true,letterpaper=true,colorlinks,bookmarks=false]{hyperref}

\dictafinalcopy 

\begin{document}

\title{From Healthy Scans to Annotated Tumors: A Tumor Fabrication Framework for 3D Brain MRI Synthesis}

\author{
Nanyu Dong$^{1}$ \quad 
Townim Chowdhury$^{1}$ \quad 
Hieu Phan$^{1}$ \quad 
Mark Jenkinson$^{1}$\\ 
Johan Verjans$^{1}$ \quad 
Zhibin Liao$^{2,1}$\thanks{Corresponding author: zhibin.liao@adelaide.edu.au}\\
\small $^{1}$Australian Institute for Machine Learning, The University of Adelaide, Australia\\
\small $^{2}$School of Computer and Mathematical Sciences, The University of Adelaide, Australia\\
{\tt\small \{nanyu.dong, townim.chowdhury, vuminhhieu.phan, mark.jenkinson,}\\
{\tt\small johan.verjans, zhibin.liao\}@adelaide.edu.au}
}

\maketitle

\begin{abstract}
The scarcity of annotated Magnetic Resonance Imaging (MRI) tumor data presents a major obstacle to accurate and automated tumor segmentation. While existing data synthesis methods offer promising solutions, they often suffer from key limitations: manual modeling is labor-intensive and requires expert knowledge. Deep generative models may be used to augment data and annotation, but they typically demand large amounts of training pairs in the first place, which is impractical in data-limited clinical settings.
In this work, we propose Tumor Fabrication (TF), a novel two-stage framework for unpaired 3D brain tumor synthesis. The framework comprises a coarse tumor synthesis process followed by a refinement process powered by a generative model. TF is fully automated and leverages only healthy image scans along with a limited amount of real annotated data to synthesize large volumes of paired synthetic data for enriching downstream supervised segmentation training.
We demonstrate that our synthetic image–label pairs used as data enrichment can significantly improve performance on downstream tumor segmentation tasks in low-data regimes, offering a scalable and reliable solution for medical image enrichment and addressing critical challenges in data scarcity for clinical AI applications.
\end{abstract}


Progress in tumor segmentation for medical imaging is significantly limited by data-related challenges~\cite{havaei2017brain,pereira2016brain}. While healthy scans are widely available, they are often underutilized. Pathological cases contain crucial signals for diagnosis, but they are rare and expensive to annotate. The difficulty of acquiring high-quality annotations, combined with strict privacy regulations that restrict data sharing~\cite{jin2019review}, further hampers the development of robust and generalizable segmentation models.

Data synthesis has emerged as a promising solution to mitigate these limitations~\cite{friedrich_frisch_cattin_2024}. Traditional synthesis techniques~\cite{prastawa2009simulation,rexilius2004framework} often rely on handcrafted manipulations, which are time-consuming, demand substantial clinical expertise, and struggle to generalize across datasets or disease types. Alternatively, deep generative models~\cite{kapoor2023multiscale,dorjsembe2022three,kim20243d} offer the potential to generate realistic data, but they typically require large training datasets and prolonged training time. Moreover, many of these models cannot produce paired synthetic labels, limiting their applicability for improving the downstream supervised segmentation tasks.

To address these challenges, we propose \textbf{Tumor Fabrication (TF)}, a novel two-stage framework for synthesizing realistic brain tumor cases directly from healthy 3D MRI scans. In the first stage, we incorporate clinical priors to simulate coarse tumor regions through region-of-interest (ROI) guided augmentation. In the second stage, we refine the synthetic samples using an adversarial generative network trained with a tailored reconstruction loss and class-wise perceptual loss. This design enhances realism within ROI while preserving anatomical fidelity in non-ROI areas.

We validate the effectiveness of TF through extensive experiments. By using our generated synthetic image–label pairs as data enrichment, the downstream segmentation model achieves state-of-the-art performance under limited real data conditions. Overall, our tumor synthesis framework provides a practical solution for leveraging healthy scans for medical data enrichment, particularly in clinical settings with limited data availability.

Our main contributions are summarized as follows:
\begin{itemize}
    \item We propose a two-stage unpaired 3D brain tumor synthesis framework (TF-Aug + TF-GAN) that generates realistic tumor image–label pairs from healthy images and a small set of unpaired real tumor data. The resulting synthetic image–label pairs effectively enhance downstream segmentation performance under limited data conditions.
    
    \item We design a region-aware GAN refinement network with a dual-head discriminator and tailored loss functions, including a novel hinge loss for anatomical preservation and a class-wise perceptual loss to enhance fine-grained tumor realism across multiple subregions.
    
    \item Extensive experiments on BraTS 2023 demonstrate that using our synthetic data pairs as enrichment significantly improves segmentation performance in data-scarce settings. It outperforms state-of-the-art data enrichment methods such as CarveMix~\cite{zhang2023carvemix} and Pix2Pix~\cite{isola2017image}.

\end{itemize}

\section{Related Work}

\noindent \textbf{Mixing-Based Synthesis.} Mixing-based methods such as Mixup~\cite{zhang2018mixupempiricalriskminimization}, CutMix~\cite{yun2019cutmixregularizationstrategytrain}, and CarveMix~\cite{zhang2023carvemix} generate synthetic samples by blending image-label pairs. While effective for improving model generalization, their success heavily depends on the diversity of annotated real data. In low-data settings, such methods may fail to capture the full complexity of tumor morphology. And simple cut-and-mix will introduce unrealistic artifacts. These observations motivate us to explore a complementary approach that combines simple cut-mix strategies with machine learning-based refinement to better incorporate anatomical and pathological realism into the synthesis process.

\noindent \textbf{Handcrafted Synthesis.} Handcrafted approaches~\cite{wang2017purine,prastawa2009simulation,clatz2005realistic,hu2023label}, such as ellipse-based modeling, deformation, and noise injection, offer controllable tumor synthesis. However, they are labor-intensive and depend on expert knowledge. Those limitations are amplified by the complexity and heterogeneity of brain tumors, making large scale realistic data generation difficult.

\noindent \textbf{Deep Generative Synthesis.} Deep generative models such as GANs~\cite{kim20243d,apoorva2021mri}, VAEs~\cite{kapoor2023multiscale}, and diffusion models~\cite{dorjsembe2022three,zhu2024generative} have shown strong potential in medical image synthesis. However, unconditional models~\cite{kim20243d,friedrich2024wdm} lack paired outputs for supervised tasks, while conditional models~\cite{apoorva2021mri,wei2019predicting} rely heavily on large paired datasets, which are often scarce in clinical domains. Our framework addresses these challenges by leveraging generative refinement with minimal annotation requirements, enabling practical use in real-world data-scarce settings.

\section{Methodology}

Let $\mathbf{x} \in \mathcal{X}$ be a single 3D MRI scan and $m \in \mathcal{M}$ be the corresponding segmentation mask, where $\mathcal{X} \in \mathbb{R}^{C \times H \times W \times D}$ and $\mathcal{M} \in \mathbb{R}^{H \times W \times D}$ denote the sets of MRI scans and segmentation masks, respectively. Here, $C$ represents the number of channels (e.g., MRI modalities), and $H$, $W$, and $D$ denote the height, width, and depth of the volume.

Given a healthy ($h$) brain MRI scan $\mathbf{x}_h$, our tumor fabrication framework aims to generate a synthetic ($s$) brain tumor MRI image $\mathbf{x}_s$ and label pairs $m_s$ by leveraging a limited set of real ($r$) tumor images $\mathbf{x}_r$ with their segmentation masks $m_r$. 
The synthetic image-label pairs are used as augmented data and directly mixed with real tumor data during downstream training to improve segmentation performance.

This framework consists of two key stages: (1) \textbf{TF-Aug (Coarse Synthetic Data Pair Generation)}, where we generate an initial tumor-affected image $\mathbf{x}_{s'}$ along with the corresponding synthetic segmentation mask $m_s$ by applying region-of-interest (ROI) augmentation on healthy MRI scans. (2) \textbf{TF-GAN (Refinement Stage)}, where the coarse synthetic image $\mathbf{x}_{s'}$ is refined into a more realistic tumor-bearing image $\mathbf{x}_s$ using our proposed TF-GAN network to enhance anatomical and visual fidelity.

\begin{figure*}[t]
\centering
\includegraphics[width=0.8\textwidth]{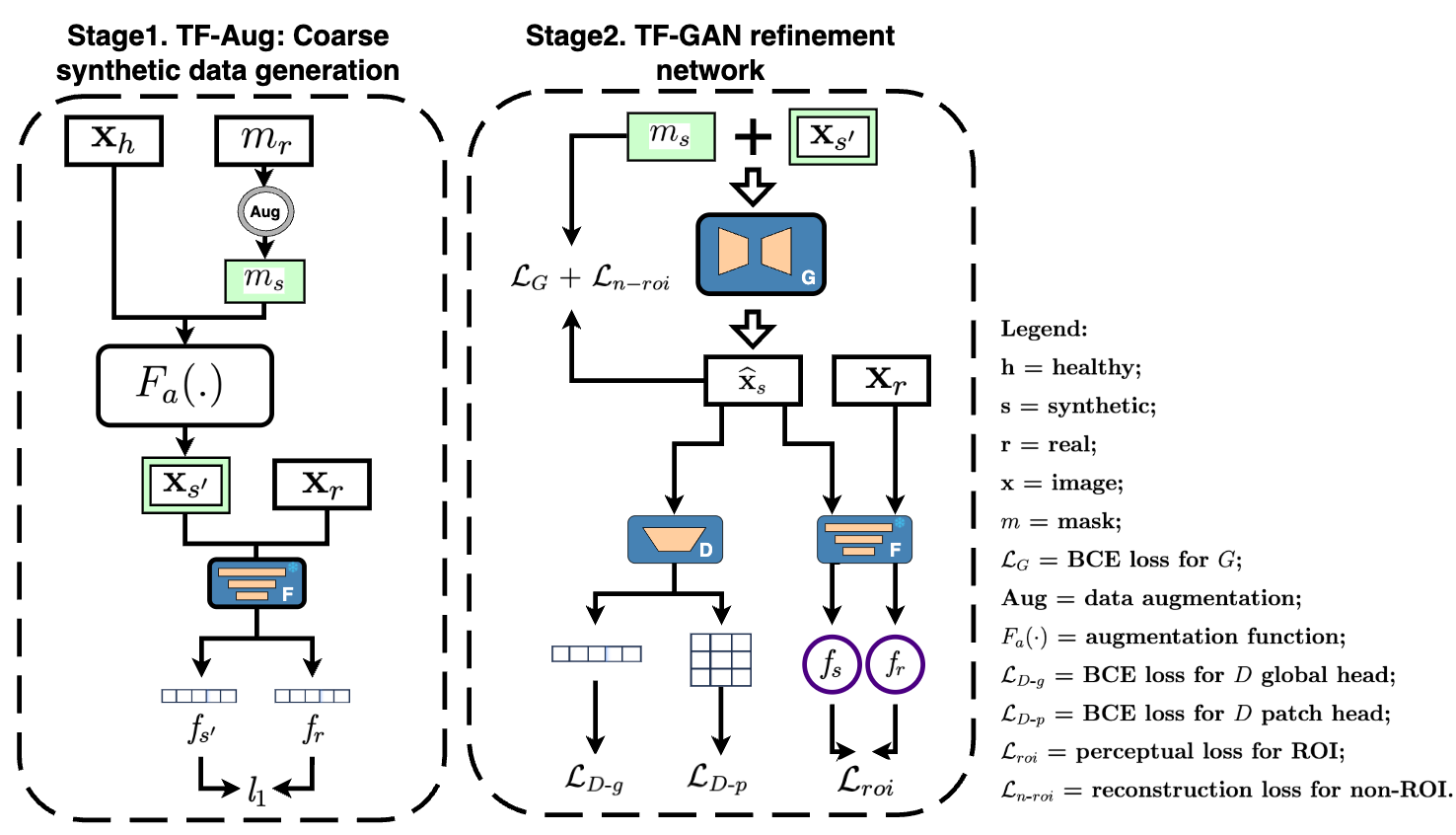}
   \caption{Overview of the proposed framework. It consists of two stages: (1) TF-Aug: generation of coarse synthetic image–label pairs $(\mathbf{x}_{s'}, m_s)$ using ROI augmentations, and (2) refinement via TF-GAN, which includes a generator $G$, a dual-head discriminator $D$, and a frozen feature extractor $F$. The training is guided by region-specific losses: $\mathcal{L}_{n\text{-}roi}$ (hinge L1 loss for non-tumor regions) and $\mathcal{L}_{roi}$ (class-wise perceptual loss for tumor subregions). 
    }
\label{fig:main_overview}
\end{figure*}

\textbf{Data Preprocessing:} All input MRI images undergo a standardized preprocessing pipeline before tumor fabrication. The preprocessing steps include: (1) Skull stripping to remove non-brain tissues \cite{hoopes2022synthstrip}. (2) Registration of healthy MRI scans to real images to align spatial characteristics, ensuring an isotropic voxel size of 1 mm. (3) Intensity normalization, where MRI intensities are rescaled to the range [-1, 1] for uniform representation. 
The overall pipeline is illustrated in Figure~\ref{fig:main_overview}.

\subsection{Stage 1: Coarse synthetic data generation}  

TF-Aug is proposed to generate coarse synthetic data pairs in stage 1. To enhance tumor shape diversity, we augment the real tumor segmentation mask $m_r$ through conventional data augmentation, including scaling, shifting, and combining multiple masks to generate variations $m_s$. Each augmented synthetic mask $m_s$ is randomly paired with a healthy brain MRI scan $\mathbf{x}_h$, and any mask regions that fall into the background area are discarded to ensure the tumor remains within the anatomical boundaries. The masked region $m_s$ is then designated as the Region of Interest (ROI) for subsequent modifications.

We perform ROI augmentation using an augmentation function:
\begin{equation}
    F_a(.) = I(B(\mathbf{x}_h, m_s)),
\end{equation}
where, $B(.)$ is the function to blur $\mathbf{x}_h$ in the masked region $m_s$ and $I(.)$ is the intensity modification function. 

\noindent \textbf{(1) ROI Blurring, $B(.)$}: A Gaussian filter with standard deviation ($\sigma$=2) is applied to the healthy tissue in the ROI area to remove high-frequency information and ensure smoother transitions between synthetic tumor area and real healthy tissue regions. 

\noindent \textbf{(2) Intensity Modification, $I(.)$}: To align the intensity distributions of synthetic and real tumors, we apply a linear transformation $I(x) = ax + b$ to each tumor subregion (ET, ED, NCR), where $a$ and $b$ are learnable parameters.

A frozen feature extractor $F$, adapted from the encoder of our baseline segmentation model~\cite{isensee2021nnu}, is used to extract embeddings $\mathbf{f}_{s'}$ and $\mathbf{f}_r$ from the synthetic image $\mathbf{x}_{s'}$ and a real image $\mathbf{x}_r$. These parameters are optimized by minimizing the L1 distance $|\mathbf{f}_{s'} - \mathbf{f}_r|$, encouraging the synthetic tumor regions to resemble real ones in intensity.

\subsection{Stage 2: Refinement of synthetic image}  

We propose a \textbf{TF-GAN refinement network} consisting of a generator $G$, a discriminator $D$, and a frozen feature extractor $F$ to refine coarse synthetic MRI scans $\mathbf{x}_{s'}$ into more realistic images $\mathbf{x}_s$.

The generator $G$ adopts a 3D U-Net-style encoder-decoder architecture with six resolution levels. To reduce checkerboard artifacts, we replaced transposed convolutions with a custom upsampling block that combines nearest-neighbor interpolation with standard convolutions.

The discriminator $D$ is a 3D convolutional network with dual output heads. A shared encoder extracts multi-scale features through five downsampling stages. The \textit{local head} follows a PatchGAN design, producing a patch-wise real/fake prediction map, while the \textit{global head} applies adaptive average pooling followed by a fully connected layer to assess the image holistically. This dual-head design allows $D$ to capture both fine-grained tumor realism and global anatomical coherence.

Additionally, we employ a frozen feature extractor $F$, derived from the encoder of a baseline segmentation model (UNet trained on real samples), to provide perceptual guidance by capturing tumor-specific feature representations. 

Adversarial training is used, where $D$ is trained to distinguish between real and refined synthetic scans, while $G$ is optimized to fool $D$. Through joint optimization, $G$ progressively improves its ability to generate realistic synthetic MRI volumes.

\subsubsection{Training Objective.}

The objective of TF-GAN is to refine the tumor regions to enhance visual and structural realism while preserving anatomical fidelity in the non-tumor areas. To achieve this, we design a set of tailored loss functions to guide the training of the refinement network. Specifically, three loss terms are employed: (1) a \textit{hinge reconstruction loss} applied to non-ROI (non-tumor) regions to enforce anatomical consistency; (2) a \textit{class-wise perceptual loss} applied within the ROI (tumor) region to encourage realistic appearance for each tumor subtype; and (3) a standard \textit{adversarial loss} computed using both the patch-based and global outputs of the dual-head discriminator to promote overall image realism.

\begin{figure}[t]
\centering
\includegraphics[width=\linewidth]{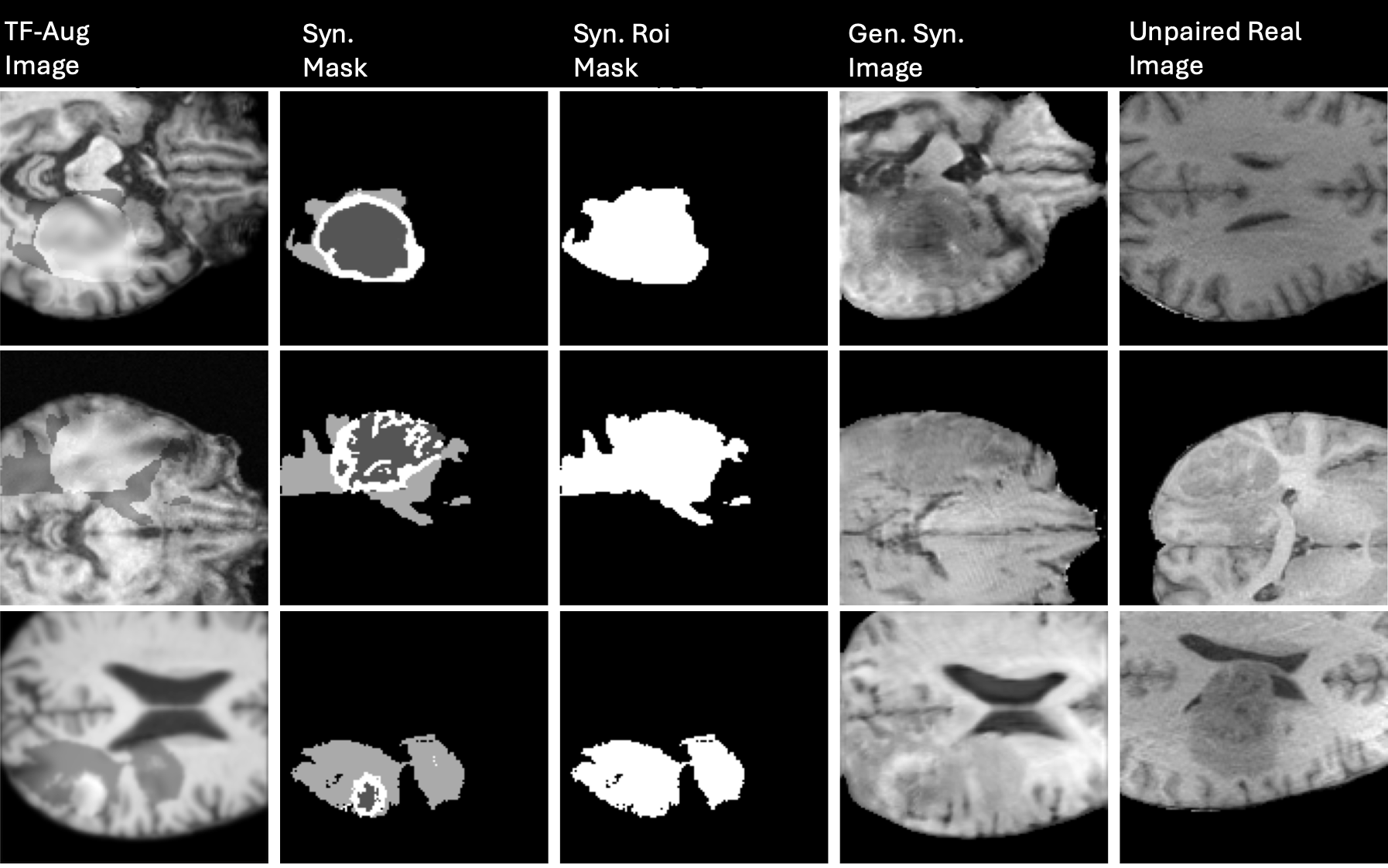}
   \caption{Sample visualization from TF-GAN refinement process. From left to right: Coarse synthetic image $\mathbf{x}_{s'}$, Synthetic mask $m_s$, ROI mask $m_{roi}$, Refined synthetic image $\mathbf{x}_s$ and an unpaired real image $\mathbf{x}_r$.}
\label{fig:training_vis}
\end{figure}

\noindent\textit{\textbf{Hinge Reconstruction Loss($\mathcal{L}_{n-roi}$).}} To preserve anatomical structure in non-tumor regions, we apply a hinge L1 loss to voxels outside the ROI, defined by the binary mask $m_{roi}$ (1 for tumor, 0 for background). Unlike standard L1 loss, the hinge loss introduces a margin to tolerate minor deviations between the refined image $\mathbf{x}_s$ and the coarse input $\mathbf{x}_{s'}$, allowing limited deformation due to tumor mass effect. The loss encourages structural consistency while maintaining flexibility near tumor boundaries:


\begin{align}
\mathcal{L}_{\text{hinge}} =\ 
\max\big(0,\ (1 - m_{\text{roi}})\odot |\mathbf{x}_{s} - \mathbf{x}_{s'}| 
- \text{margin}_{\text{n-roi}}\big),
\label{eq:hinge}
\end{align}

where $\mathbf{x}_{s}$ and $\mathbf{x}_{s'}$ represent the generated refined and coarse synthetic MRI scans, respectively, and $\odot$ denotes element-wise multiplication. Eq. \ref{eq:hinge}  Visualization during training can be found in Fig. \ref{fig:training_vis}.


\noindent\textit{\textbf{Class-wise Perceptual Loss($\mathcal{L}_{roi}$)}} .
To enhance the visual fidelity of tumor sub-regions and maintain semantic consistency, we design a class-wise perceptual loss over the ROI. Specifically, we use a frozen feature extractor $F$ to produce multi-scale feature maps. Only the first three layers were used due to their higher spatial resolution and ability to capture low-level appearance details.
Given the generated image and a random real image, we compute their respective feature representations via $F$. For each tumor class, we extract class-specific feature vectors by applying the corresponding segmentation masks at each feature resolution. The class-wise features are aggregated using masked global average pooling. The perceptual loss is then computed as the mean L1 distance between real and generated features, averaged across layers and classes. This loss encourages the synthesized tumor regions to exhibit class-consistent textural and structural patterns that align with real samples. The overall class-wise perceptual loss is defined as:


\begin{align}
    \mathcal{L}_{\text{percep}} &= \sum_{i=1}^{L} \sum_{c=1}^{C} \left| f_r^{(i,c)} - f_s^{(i,c)} \right|,
    \label{eq:perceptual}
\end{align}

where $L$ is the number of feature layers used, $C$ is the number of tumor classes (excluding background), $f_r^{(i,c)}$ and $f_s^{(i,c)}$ denote the global average pooled features of class $c$ at layer $i$ for the real and synthetic images respectively.


\noindent\textit{\textbf{Adversarial Loss.}} The generator $G$ and discriminator $D$ are trained in an adversarial way by solving the Minimax objective $\arg\min_G\,\arg\max_D\, \mathcal{L}_{adv}$~\cite{goodfellow2020generative}. 
Given the dual-head design of $D$, which outputs both patch-level and global scores, we define the total adversarial loss as the sum of two components $\mathcal{L}_{adv}^{patch}$ and $\mathcal{L}_{adv}^{global}$, and for each one, the adversarial loss is defined as:

\begin{align}
\mathcal{L}_{adv} =\ & \mathbb{E}_{\mathbf{x}_r \sim p(\mathbf{x}_r)}[\log(D(\mathbf{x}, m))] \nonumber \\
& + \mathbb{E}_{\mathbf{x}_{s'} \sim p(\mathbf{x}_{s'})}[\log(1 - D(G(\mathbf{x}_{s'}), m))] ,\label{eq:adv}
\end{align}

\noindent\textit{\textbf{Overall Objective.}} The total training objective for the refinement network combines adversarial supervision, structural preservation, and perceptual alignment. Specifically, we incorporate both patch-wise and global adversarial losses, a hinge-based reconstruction loss for non-tumor regions, and a class-wise perceptual loss for tumor regions. The overall objective is defined as:

\begin{align}
\mathcal{L}_{\text{total}} =\ 
& \lambda_{a} \cdot \mathcal{L}_{\text{adv}}^{\text{patch}} +
  \lambda_{b} \cdot \mathcal{L}_{\text{adv}}^{\text{global}} \nonumber \\
& + \lambda_{c} \cdot \mathcal{L}_{\text{hinge}} +
  \lambda_{d} \cdot \mathcal{L}_{\text{percep}},
\label{eq:total_loss}
\end{align}
\noindent
where $\lambda_{a}$, $\lambda_{b}$, $\lambda_{c}$, and $\lambda_{d}$ are weighting coefficients that balance the contribution of each loss term. Specifically, we set $\lambda_{a} = 10$ and $\lambda_{b} = 1$ to account for the relatively smaller magnitude of the patch-level adversarial loss, ensuring it remains on a comparable scale with the global loss. The reconstruction loss weight $\lambda_{c}$ is initialized at $10$ and linearly decreased to $1$ over training. This dynamic schedule emphasizes structural consistency in the early training stages, providing stronger gradients to guide global appearance learning. Finally, the perceptual loss weight $\lambda_{d}$ is fixed at $1$ throughout training.

\section{Experiments and Results}

\subsection{Experimental Settings}

\subsubsection{Datasets.}  
We utilize T1-weighted MRI scans from the IXI~\cite{IXIDataset} and PPMI~\cite{PPMI} datasets as unlabelled healthy source data for synthetic tumor generation. For real tumor segmentation, we adopt the BraTS 2023~\cite{kazerooni2024braintumorsegmentationbrats} adult glioma task. To simulate a data-scarce clinical scenario, we randomly select 100 T1-weighted MRI scans with corresponding segmentation labels for training, including the generation and downstream segmentation. An additional 200 labeled cases from BraTS 2023 are held out for validation.

\subsubsection{Implementation Details.}  
All experiments were conducted using PyTorch on two NVIDIA RTX A6000 GPUs (48 GB each). 

Stage 1 (TF-Aug) was trained for 200 epochs using stochastic gradient descent (SGD) with a learning rate of $1 \times 10^{-2}$.

Stage 2 (TF-GAN) involved adversarial training with random cropping of $128 \times 128 \times 128$ patches and standard augmentations. The one-hot encoded synthetic mask $m_s$ was concatenated with the image along the channel dimension as input. We used the Adam optimizer~\cite{kingma2014adam} with $(\beta_1, \beta_2) = (0.5, 0.999)$. The learning rates were set to $2 \times 10^{-4}$ for the generator and $1 \times 10^{-4}$ for the discriminator, both linearly decayed over 200 epochs.

Downstream segmentation was performed using the nnU-Net framework~\cite{isensee2021nnu}, configured with a six-stage encoder–decoder architecture and input patch size of $128 \times 128 \times 128$. We adopted the default nnU-Net settings, including optimizer parameters, learning rate schedules, data augmentations, and post-processing. All experiments were trained for 100 epochs.


\subsection{Results}

\subsubsection{Compared Methods.}  
We compare our method (\textbf{TF}), which directly utilizes synthetic image-label pairs as data enrichment for downstream segmentation, against several representative data synthesis approaches. All compared methods are adapted to 3D where applicable. All GAN-based methods were trained for 200 epochs.

(1) CraveMix~\cite{zhang2023carvemix}: A cut-and-mix based data augmentation strategy that carves the tumor region along with a surrounding contextual margin from a real tumor-bearing scan and blends it into another real tumor-bearing scan. The corresponding segmentation masks are updated accordingly.

(2) Pix2Pix~\cite{isola2017image}: A conditional GAN model designed for paired data. We construct input-target pairs by extracting the tumor segmentation masks from real scans and added an extra label stands for non-background brain area(exclusive from the tumor masks). These pairs are used to train a 3D Pix2Pix model.

\subsubsection{Metric Definition.}  
\noindent \textit{Dice Score} is used as the evaluation metric for the multiclass tumor segmentation task. Following the standard BraTS protocol, we compute the Dice score separately for three clinically relevant subregions: enhanced tumor (ET), tumor core (TC), and whole tumor (WT), as well as the mean Dice score across these three classes. All metrics are computed using the nnU-Net framework with a consistent post-processing pipeline across all experiments.

The Dice score quantifies the volumetric overlap between predicted segmentation $\hat{Y}$ and ground truth $Y$, and is defined as:

\begin{equation}
\text{Dice} = \frac{2 \sum_{i=1}^{I} Y_i \hat{Y}_i}{\sum_{i=1}^{I} Y_i + \sum_{i=1}^{I} \hat{Y}_i},
\label{eq:dice}
\end{equation}
where $Y_i$ and $\hat{Y}_i$ denote the ground truth and predicted binary labels for voxel $i$, respectively. A higher Dice score indicates better segmentation performance.

\subsubsection{Quantitative Comparisons.}  
Table~\ref{tab:main_table}presents the segmentation performance of our proposed full method TF and several comparison synthesis approaches on the BraTS 2023 dataset~\cite{kazerooni2024braintumorsegmentationbrats}, using only T1-weighted scans. For each method, the training set consists of 100 real image-label pairs mixed with 100 synthetic image-label pairs generated by the respective approach. All experiments are conducted using nnU-Net with consistent augmentations and post-processing. Results are averaged over three independent runs with different random seeds.

Our full method TF, achieves the highest mean Dice score (67.77\%), along with the best performance on ET (51.75\%) and TC (69.98\%), highlighting the effectiveness of our refinement stage in improving tumor boundary realism and fine-grained structural fidelity. While its WT score (81.59\%) is competitive, it is slightly lower than CarveMix, suggesting that refinement may focus more on localized tumor details rather than broader tumor structures like edema.

Importantly, our method achieves a statistically significant improvement over the baseline, with a two-tailed t-test yielding $p < 0.05$. It also demonstrates lower variance across runs, indicating greater robustness. Although the mean Dice gain over the baseline is 1.2\%, this improvement is consistent and clinically meaningful. In contrast, CarveMix and Pix2Pix provide only modest and less stable improvements. These results highlight the effectiveness of our structured two-stage refinement in generating high-quality synthetic data that translates to reliable segmentation gains.

\begin{table*}[!htbp]
    \centering
    \resizebox{0.75\textwidth}{!}{
    \begin{tabular}{c|c|c|c|c|c|c}
    \toprule
    Method  & ET(\%)$\uparrow$ & TC(\%)$\uparrow$ & WT(\%)$\uparrow$ & Mean(\%)$\uparrow$ & Mean Diff(\%)$\uparrow$ & p-value$\downarrow$\\
    \hline\hline
    \textit{Baseline(nnUnet)}& 49.76 $\pm$0.69  & 68.45$\pm$0.64  & 81.50$\pm$0.41  & 66.57$\pm$0.52  & - & -\\
    \hline
    CarveMix& 49.81$\pm$0.25  & 68.55$\pm$0.41  &\textbf{ 81.80$\pm$0.27 } & 66.72$\pm$0.25 & +0.15 & 0.67 \\
    \hline
    Pix2pix& 50.12$\pm$0.61  & 67.89$\pm$0.38  & 80.66$\pm$0.34  & 66.23$\pm$0.44 & -0.34 & 0.43\\
    \hline
    
    \midrule
    TF(ours) & \textbf{51.75$\pm$0.21}  & \textbf{69.98$\pm$0.09}  & 81.59$\pm$0.07 & \textbf{67.77$\pm$0.08} & \bf{+1.20} & \textbf{0.02}\\
    \bottomrule
    \end{tabular}
    }
    \caption{Quantitative results on the downstream segmentation task. ET, TC, and WT represent the enhancing tumor, tumor core, and whole tumor regions, respectively. The Mean column reports the average Dice score across these three tumor subregions. The Mean Diff column highlights the Mean column difference between each compared method and the Baseline. The p-value column presents the results of a two-tailed t-test, where a value below 0.05 indicates a statistically significant improvement over the Baseline. Results are presented as mean and standard deviation over three independent runs with different random seeds. The best performance in each column is highlighted in bold.}
    \label{tab:main_table}
\end{table*}

\subsubsection{Qualitative Comparisons.}
\begin{figure*}[t]
\centering
\includegraphics[width=0.60\linewidth]{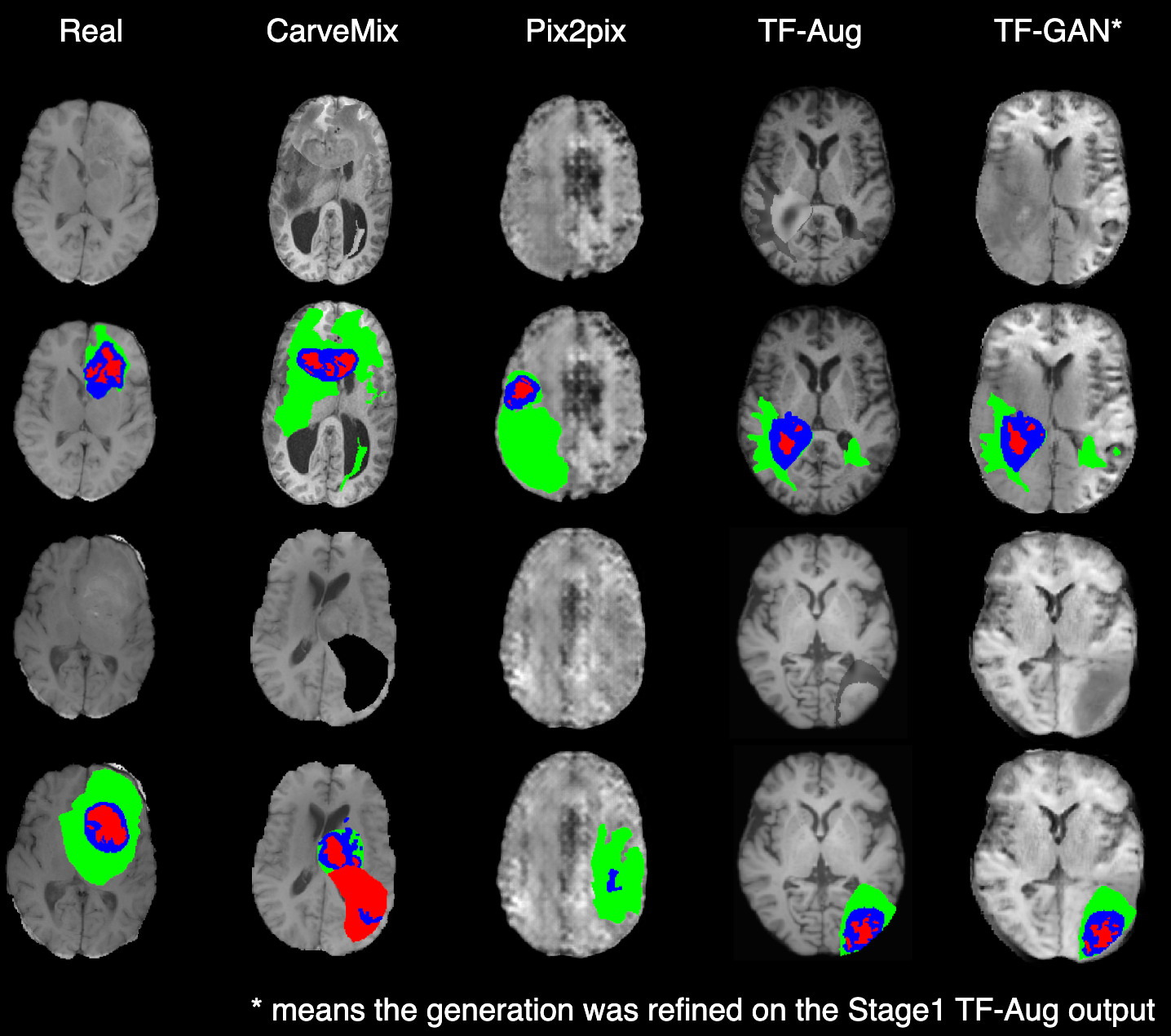}
\caption{Qualitative comparison between our method and other baselines. For each method, two representative samples are shown, including the synthetic image and its corresponding segmentation labels. For Tf-Aug and TF-GAN, identical input masks are used to enable direct visual comparison of refinement quality.}
\label{fig:qualitative}
\end{figure*}

Figure~\ref{fig:qualitative} presents visual comparisons of representative samples generated by different methods. The first column shows real T1-weighted MRI scans, illustrating the inherent difficulty of the task: tumor subregions such as edema, tumor core, and enhancing tumor often exhibit blurred boundaries, irregular shapes, and heterogeneous appearances. These characteristics underscore the importance of generating anatomically realistic and structurally diverse tumor regions.

CarveMix\cite{zhang2023carvemix}, which performs cut-and-mix augmentation, introduces visible artifacts. Tumor regions often exhibit unnatural transitions or label inconsistencies. For example, edema labels appearing inside the tumor core. Intensity mismatches between different scans can also produce visually jarring features, such as sudden dark spots or patchy edges.

Pix2Pix\cite{isola2017image}, a paired label-to-image translation model, generates overly smooth and blurry outputs. This is likely due to the limited representational capacity of the input label maps, which lack fine-grained texture or structural information, resulting in less realistic synthetic scans.

In contrast, the last two columns show results from using TF-Aug alone versus TF-Aug followed by TF-GAN refinement. The same input samples are listed for direct comparison. TF-Aug alone, despite relying only on intensity modification and ROI blurring, benefits from the high anatomical fidelity of healthy brain scans and produces sharper structures. However, the tumors appear overly clean and unrealistic, with sharp boundaries and little blending with surrounding tissues.

Refinement through TF-GAN significantly enhances the realism of synthetic tumors. It smooths tumor edges and integrates them more naturally into surrounding tissues. While minor square-shaped artifacts occasionally appear, likely due to sliding-window inference, they have minimal visual impact. TF-GAN produces clearer and more detailed outputs, particularly in non-tumor regions, and shows improved tumor boundary blending. These results underscore the effectiveness of TF-GAN in applying tumor-aware refinements while preserving global anatomical coherence.

Despite the effectiveness of our two-stage framework, some limitations persist. The synthesized edema regions often appear overly diffuse and anatomically ambiguous, which may explain the relatively modest Dice improvements for the whole tumor (WT) region, where edema dominates. Additionally, the generated tumors lack realistic mass effect, such as tissue displacement, resulting in more symmetric and less deformed brain structures compared to real glioma scans. Future work will focus on better modeling edema characteristics and mass effect deformations to enhance anatomical realism.

\subsubsection{Ablation Study on Synthetic Data Volume and Refinement.}  
We conduct an ablation study to assess two factors: (1) the impact of increasing the volume of synthetic data, and (2) the effect of the proposed TF-GAN refinement network over TF-Aug coarse synthesis alone.

\begin{table}[!htbp]
    \centering
    \resizebox{0.9\linewidth}{!}{
    \begin{tabular}{c|c|c|c|c|c}
    \toprule
    Exp. & ET(\%)$\uparrow$ & TC(\%)$\uparrow$ & WT(\%)$\uparrow$ & Mean(\%)$\uparrow$ & Mean Diff(\%)$\uparrow$ \\
    \hline\hline
    \textit{Baseline} & 49.76 $\pm$0.69  & 68.45$\pm$0.64  & 81.50$\pm$0.41  & 66.57$\pm$0.52 & - \\
    \hline
    Syn50 & 50.24$\pm$0.23 & 68.72$\pm$0.23  &  81.68$\pm$0.37 & 66.88$\pm$0.28 & +0.31 \\
    \hline
    Syn100 &51.75$\pm$0.21  & 69.98$\pm$0.09 & 81.59$\pm$0.07 & 67.77$\pm$0.08 &+1.20\\
    \hline
    Syn200 &  51.36$\pm$0.76 & 69.75$\pm$0.60  &  81.56$\pm$0.10 & 67.56$\pm$0.49 & +1.18\\
    \hline
    Syn300 &  \textbf{52.16$\pm$0.78} & \textbf{70.61$\pm$0.52}  &  \textbf{81.99$\pm$0.09} & \textbf{68.25$\pm$0.46} & \textbf{+1.68}  \\
    \bottomrule
    \end{tabular}
    }
    \caption{Ablation study evaluating the impact of different volumes of synthetic data added to a fixed set of 100 real training samples. ET, TC, and WT represent the mean Dice scores for the enhancing tumor, tumor core, and whole tumor regions, respectively. The Mean column reports the average Dice score across these three tumor subregions. The Mean Diff column highlights the Mean column difference between each compared method and the Baseline. The best performance in each column is highlighted in bold.}
    \label{tab:ablation}
\end{table}

First, with a fixed set of 100 real training samples, we incrementally add 50, 100, 200, and 300 synthetic image-label pairs generated by TF-GAN. As shown in Table~\ref{tab:ablation}, Segmentation performance shows a consistent upward trend with increasing synthetic data, as reflected in the rising mean Dice scores, though individual subregion results may exhibit minor fluctuations, demonstrating the benefit of our method in limited data regimes. However, the performance gains gradually plateau, indicating diminishing marginal returns. This trend suggests that while additional synthetic data continues to enhance generalization, its incremental contribution reduces as the model saturates its learning capacity. Importantly, even when synthetic data significantly outnumbers real data, no performance degradation is observed, indicating that our refined samples preserve high domain fidelity, successfully bridging the gap between healthy and tumor domains.

\begin{table}[!htbp]
    \centering
    \resizebox{0.9\linewidth}{!}{
    \begin{tabular}{c|c|c|c|c}
    \toprule
    Exp. & ET(\%)$\uparrow$ & TC(\%)$\uparrow$ & WT(\%)$\uparrow$ & Mean(\%)$\uparrow$ \\
    \hline\hline
    \textit{Baseline} & 49.76 $\pm$0.69  & 68.45$\pm$0.64  & 81.50$\pm$0.41  & 66.57$\pm$0.52  \\
    \hline
    TF-Aug& 50.07$\pm$0.55  & 68.03$\pm$0.36  & 80.83$\pm$0.49  & 66.31$\pm$0.36  \\
    \hline
    TF-Aug+TF-GAN &  \textbf{51.75$\pm$0.21}  & \textbf{69.98$\pm$0.09}  & \textbf{81.59$\pm$0.07} & \textbf{67.77$\pm$0.08}  \\ 
    \bottomrule
    \end{tabular}
    }
    \caption{Ablation study evaluating the impact of using TF-Aug alone versus TF-Aug followed by TF-GAN refinement. ET, TC, and WT represent the mean Dice scores for the enhancing tumor, tumor core, and whole tumor regions, respectively. The Mean column reports the average Dice score across these three tumor subregions. Results are presented as mean and standard deviation over three independent runs with different random seeds. The best performance in each column is highlighted in bold.}
    \label{tab:ablation2}
\end{table}

Second, we compare the performance of using TF-Aug alone versus TF-Aug followed by TF-GAN refinement. As already reflected in the Table~\ref{tab:ablation2}, adding the refinement stage yields consistent gains across all metrics. 
Although TF-Aug alone yields slightly underperforms in some regions, this is expected due to the unrefined appearance and domain gap of the coarse synthetic images. However, TF-Aug plays a critical foundational role by providing structurally plausible tumor–label pairs, which are essential for any subsequent refinement. This further validates that the two-stage synthesis pipeline enhances the realism and effectiveness of the generated samples, beyond what is achievable by coarse generation alone.

\section{Conclusion and Future Work}

We proposed Tumor Fabrication (TF), a novel two-stage framework that generates realistic and anatomically consistent brain tumor image–label pairs by leveraging healthy MRI scans and a small amount of annotated tumor data. Combining clinically inspired augmentation (TF-Aug) with a generative refinement module (TF-GAN), TF produces high-quality synthetic data that significantly improves segmentation performance in data-scarce scenarios.

Our method provides a practical pathway for transforming abundant healthy scans into useful pathological training data, demonstrating the untapped potential of healthy images to enrich supervised learning when annotated pathology is limited.

TF is both flexible and extensible: TF-Aug enables potential control over tumor attributes (\eg, location, size, shape), while TF-GAN is capable of supporting modality translation and anatomically coherent refinements. Though not yet fully explored, these aspects highlight the framework's adaptability.

Future work will focus on improving synthesis realism and unlocking TF’s customization capabilities to support broader clinical applications, ultimately advancing robust and scalable AI solutions in medical imaging.

\vspace{15mm}

{\small
\bibliographystyle{ieee}
\bibliography{egbib}
}

\end{document}